\pdfoutput=1

\documentclass[11pt]{article}

\usepackage[preprint]{acl}

\usepackage{times}
\usepackage{latexsym}

\usepackage{colortbl}
\usepackage{graphics}

\usepackage[T1]{fontenc}

\usepackage[utf8]{inputenc}

\usepackage{microtype}

\usepackage{inconsolata}

\usepackage{graphicx}

\usepackage{xspace}
\usepackage{booktabs}
\usepackage{multirow}
\usepackage{amssymb}
\usepackage{xcolor}
\usepackage{subcaption}
\usepackage{nameref}
\usepackage{enumitem}
\usepackage{empheq, bm}
\usepackage{listings}

\usepackage{todonotes}
\usepackage{menukeys}
\usepackage[all]{nowidow}

\definecolor{codegreen}{rgb}{0,0.6,0}
\definecolor{codegray}{rgb}{0.5,0.5,0.5}
\definecolor{codepurple}{rgb}{0.58,0,0.82}
\definecolor{backcolour}{rgb}{0.95,0.95,0.92}
 
\lstdefinestyle{mystyle}{
    backgroundcolor=\color{backcolour},   
    commentstyle=\color{codegreen},
    keywordstyle=\color{magenta},
    numberstyle=\tiny\color{codegray},
    stringstyle=\color{codepurple},
    basicstyle=\ttfamily\scriptsize,
    breakatwhitespace=false,         
    breaklines=true,                 
    captionpos=b,                    
    keepspaces=true,                 
    numbers=left,                    
    numbersep=5pt,                  
    showspaces=false,                
    showstringspaces=false,
    showtabs=false,                  
    tabsize=2
}
 
\usepackage{lipsum}
\usepackage{comment}

\def\tabref#1{Table \ref{#1}}
\def\figref#1{Figure \ref{#1}}

\def\eqref#1{Eq (\ref{#1})}

\definecolor{backcolour}{rgb}{0.95,0.95,0.92}

\title{Lighthouse: A User-Friendly Library for Reproducible \\ Video Moment Retrieval and Highlight Detection}

\author{Taichi Nishimura \hspace{1.5em} Shota Nakada \hspace{1.5em} Hokuto Munakata \hspace{1.5em} Tatsuya Komatsu \\
  LY Corporation \\ \texttt{\{tainishi,shota.nakada,hokuto.munakata,komatsu.tatsuya\}@lycorp.co.jp}
}

\begin{document}
\maketitle
\begin{abstract}
We propose Lighthouse, a user-friendly library for reproducible video moment retrieval and highlight detection (MR-HD). Although researchers proposed various MR-HD approaches, the research community holds two main issues. The first is a lack of comprehensive and reproducible experiments across various methods, datasets, and video-text features.
This is because no unified training and evaluation codebase covers multiple settings. The second is user-unfriendly design. Because previous works use different libraries, researchers set up individual environments. In addition, most works release only the training codes, requiring users to implement the whole inference process of MR-HD. Lighthouse addresses these issues by implementing a unified reproducible codebase that includes six models, three features, and five datasets. In addition, it provides an inference API and web demo to make these methods easily accessible for researchers and developers. Our experiments demonstrate that Lighthouse generally reproduces the reported scores in the reference papers. The code is available at \url{https://github.com/line/lighthouse}.
\end{abstract}

\section{Introduction}

\begin{figure}[t]
  \centering
  \includegraphics[width=\linewidth]{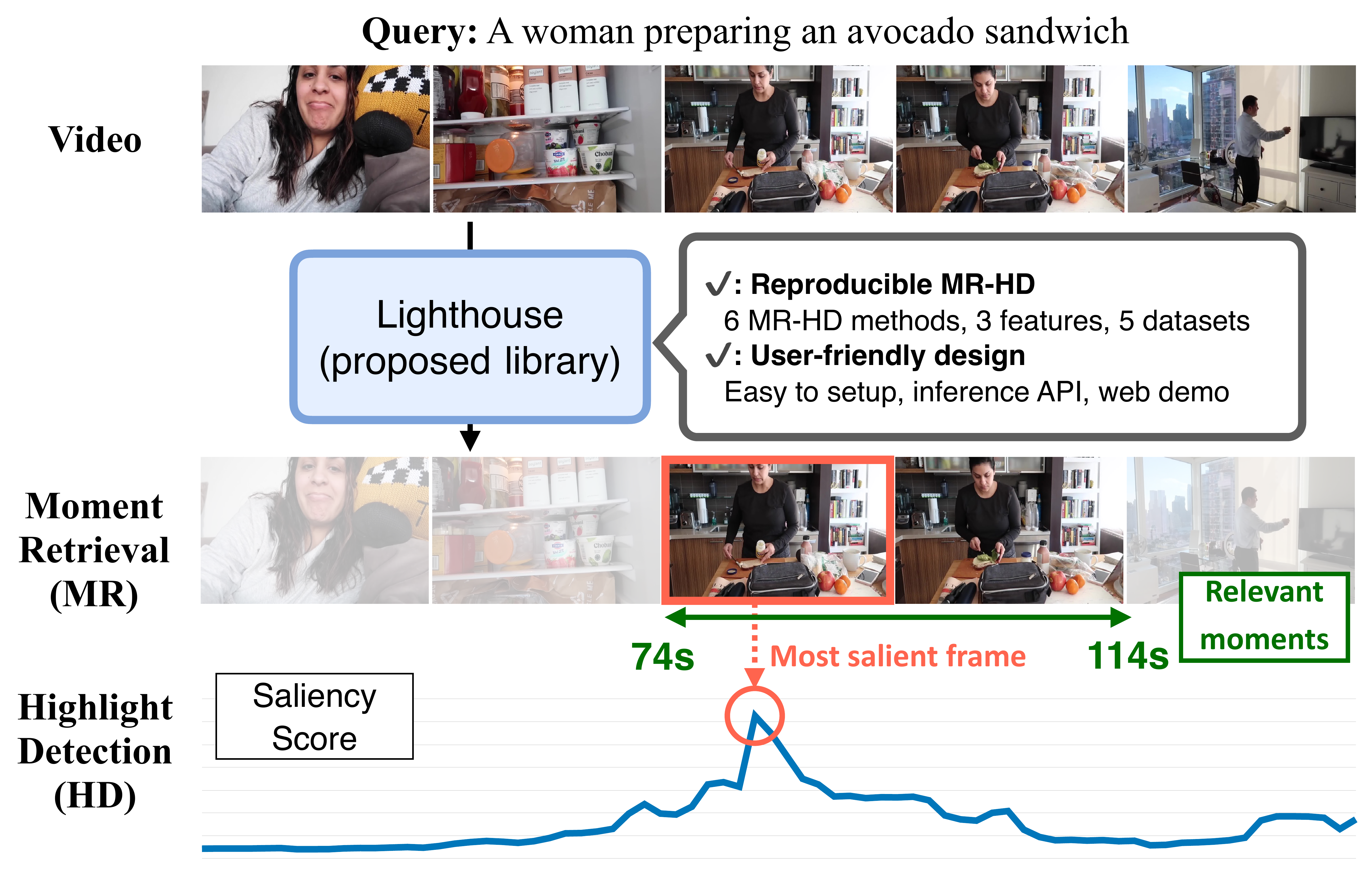}
  \caption{Overview of MR-HD and Lighthouse. Given a video and query, the model predicts relevant moments for MR and saliency scores for HD. Lighthouse achieves reproducible MR-HD by supporting multiple settings. In addition, it aims at a user-friendly design with an easy-to-setup environment, inference API, and web demo.}
  \label{fig:mr_hd}
\end{figure}

With the rapid advance of digital platforms, videos become ubiquitous and popular on the web. Although they offer rich, informative, and entertaining content, watching entire videos can be time-consuming. Hence, there is a high demand for multimodal tools that enable users to quickly find specific moments within videos and browse through highlights in the moments from natural language queries. The former is called moment retrieval (MR) and the latter is called highlight detection (HD).
Given a video and a language query, MR retrieves relevant moments (start and end timestamps), and HD detects highlighted frames within these moments by calculating saliency scores representing frame-level highlightness (\figref{fig:mr_hd}). Note that HD calculates saliency scores for all frames in the video, but the frames with the highest saliency scores are detected within the moments.

Although MR and HD share common characteristics, such as learning the similarity between input queries and video frames, they were separately treated due to the lack of annotations supporting both tasks~\citep{zhang2020acl,song2015cvpr}.
To address this, \citet{lei2021neurips} proposed the QVHighlights dataset comprising videos, language queries, and moment/highlight annotations, enabling researchers to tackle both tasks simultaneously. We refer to this unified task of MR and HD as \textit{MR-HD} to distinguish it from the individual tasks of MR and HD. Based on this dataset, various approaches have been proposed to perform MR-HD. Note that most methods are applicable for single tasks of either MR or HD as well as MR-HD.

\begin{table*}[t]
\centering
\scalebox{0.66}{
\begin{tabular}{lcccccccc}
\toprule
 & \textbf{MR-HD} & \multicolumn{3}{c}{\textbf{MR}} & \textbf{HD} & & \\ \cmidrule(lr){2-2} \cmidrule(lr){3-5} \cmidrule(lr){6-6} & \textbf{QVHighlights} & \textbf{ActivityNet Captions} & \textbf{Charades-STA} & \textbf{TaCoS} & \textbf{TVSum} & \textbf{Features} & \textbf{API?} & \textbf{Web demo?} \\ \midrule
Moment DETR \cite{lei2021neurips} & \checkmark &  &  &  &  & C+S &  &  \\
QD-DETR \cite{Moon_2023_CVPR} & \checkmark &  &  &  & \checkmark & C+S &  &  \\
EaTR \cite{Jang_2023_ICCV} & \checkmark & \checkmark & \checkmark &  &  & C+S &  &  \\
TR-DETR \cite{sum2024aaai} & \checkmark &  &  &  & \checkmark & C+S &  &  \\
UVCOM \cite{Xiao_2024_CVPR} & \checkmark &  & \checkmark &  &  & C+S &  &  \\
CG-DETR \cite{Moon2024arxiv} & \checkmark &  & \checkmark & \checkmark & \checkmark & C+S+V+G &  &  \\ \midrule
\textbf{Lighthouse (ours)} & \checkmark & \checkmark & \checkmark & \checkmark & \checkmark & C+S+R+G & \checkmark & \checkmark \\ \bottomrule
\end{tabular}
}
\caption{Comparison of Lighthouse and existing publicly available MR-HD repositories. C, S, V, R, and G in the ``Features'' column represent CLIP~\cite{radford2021icml}, Slowfast~\cite{Feichtenhofer_2019_ICCV}, VGGNet16~\cite{simonyan2014arxiv}, ResNet152~\cite{he2016cvpr}, and GloVe~\cite{glove}, respectively.}
\label{tab:comparison_repo}
\end{table*}

Despite the rapid development of MR-HD, the research community holds two issues. The first is a lack of comprehensive and reproducible experiments across various methods, datasets, and features. This is because there is no unified training and evaluation codebase covering multiple settings.
While previous work reported scores for their methods on individual tasks for MR, HD, and MR-HD, researchers release their code only for QVHighlights, without necessarily providing training codes for other datasets. In addition, datasets and features are not standardized. Researchers use different MR and HD datasets to demonstrate their approach's effectiveness (\tabref{tab:comparison_repo}). Hence, to fully reproduce experiments, researchers should set up individual environments and write additional code, ranging from video-text feature extraction preprocessing to modifications to the training and evaluation codes. This is time-consuming and cumbersome.

The second is user-unfriendly design. Because previous works use different libraries for their method, MR-HD researchers should set up individual environments. In addition, most previous works release only training codes, requiring users to implement the whole inference process of MR-HD and apply it to their videos. This includes frame extraction from videos, video-text feature extraction, and forwarding them into the trained model. Implementing all of these steps accurately is challenging for developers who are interested in MR-HD but lack expertise in video-text processing.

Our goal is to address these issues and foster the MR-HD research community.
To this end, we propose \textit{\textbf{Lighthouse}}, a user-friendly library for reproducible MR-HD. Lighthouse unifies training and evaluation codes to support six recent MR-HD methods, three features, and five datasets for MR-HD, MR, and HD, resolving the reproducibility issue.
While this results in 90 possible configurations (6 methods $\times$ 3 features $\times$ 5 datasets), the configuration files are written in YAML format, allowing researchers to easily reproduce experiments by specifying the necessary file. Our experiments demonstrate that Lighthouse mostly reproduces the original experiments in the referenced six papers. In addition, to resolve the user-unfriendliness, Lighthouse provides an inference API and web demo. The inference API covers the entire MR-HD process and provides users with easy-to-use code for MR-HD. The web demo, built upon the API, enables users to confirm the results visually. The codes are under the Apache 2.0 license.

\section{Highlights of Lighthouse}
\tabref{tab:comparison_repo} shows a comparison of Lighthouse and public MR-HD repositories. We describe them in terms of reproducibility and user-friendly design.

\subsection{Reproducibility}

\begin{figure}[t]
    \centering
    \includegraphics[width=\linewidth]{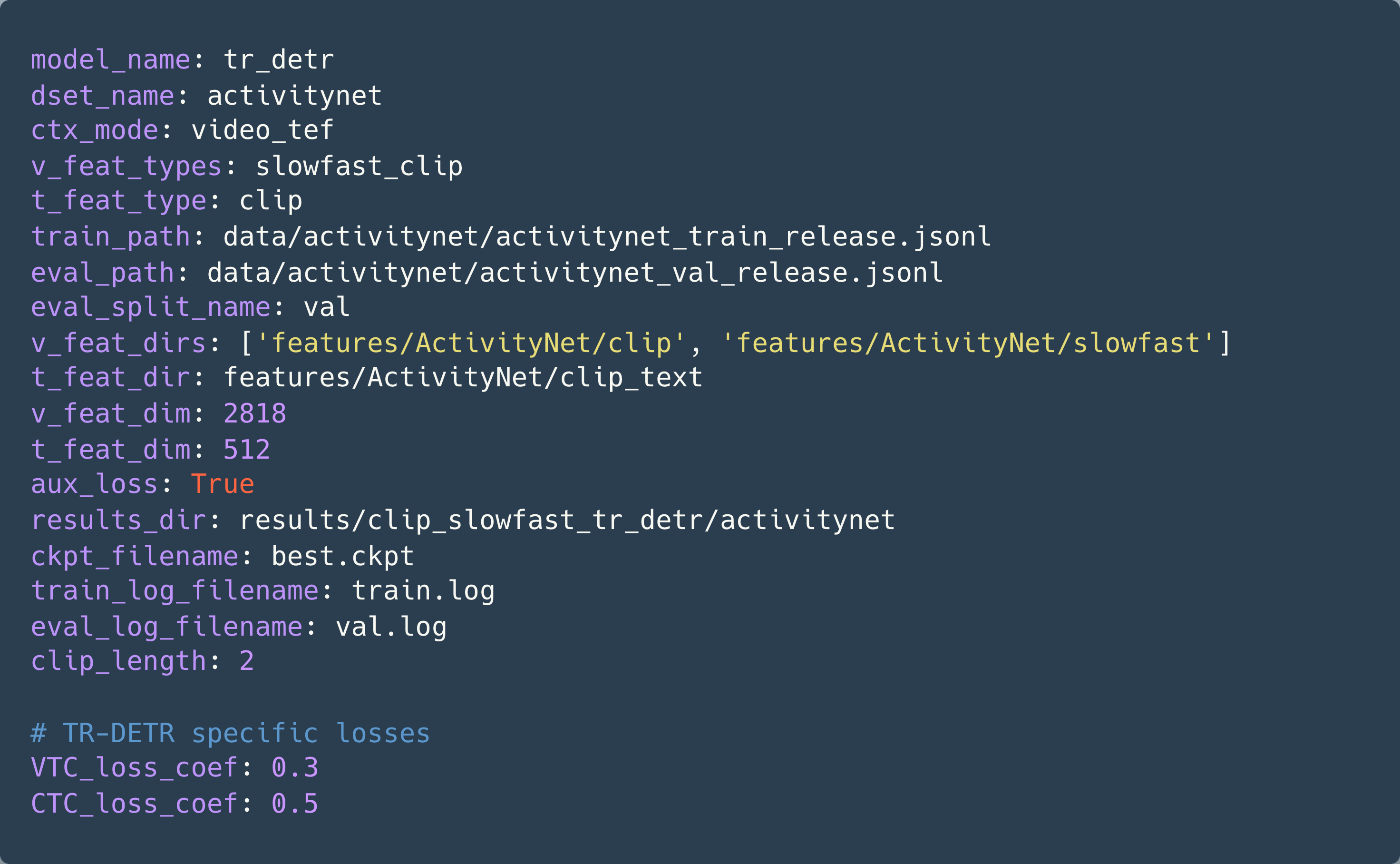}
    \caption{YAML configuration example.}
    \label{fig:yaml_config}
\end{figure}

\noindent\textbf{Support for multiple methods, datasets, and features:}
As shown in \tabref{tab:comparison_repo}, previous works support different datasets and features for MR and HD tasks. Lighthouse supports all of them by integrating all these MR-HD methods, features, and datasets into a single codebase. We extract video-text features from all datasets, train models using these features, and release reproducible code along with the features and pre-trained weights. This significantly reduces the effort required to write additional code for conducting experiments across multiple settings.

\noindent\textbf{Reproducible training and evaluation:}
Lighthouse enables researchers to reproduce the training process with a single Python command by specifying the configuration files, where hyper-parameters are written in YAML format (\figref{fig:yaml_config}).
The Lighthouse users can easily test different hyper-parameters by modifying these files.
We release all of the files used or generated during experiments, including video-text features, trained weights, and logs during the training. Therefore, to reproduce the experiments, researchers can obtain the same results by downloading the necessary files and running a single Python evaluation command with the trained weights.

\subsection{User-friendly design}

\noindent\textbf{Easy to set up:} Lighthouse allows researchers and developers to install it easily with ``\texttt{pip install .}'' after cloning the repository. Because the libraries used in previous work vary between repositories, researchers need to set up individual environments by cloning each repository and installing the dependency libraries. Lighthouse streamlines this process by summarizing the necessary libraries and carefully removing any unnecessary ones that are imported but not used in the codebase.

\begin{figure}
\begin{lstlisting}[language=Python, numbers=left, caption={Example usage of the inference API.}, label={listing:util}]
import torch
from Lighthouse.models import CGDETRPredictor

device = 'cuda' if torch.cuda.is_available() else 'cpu'

# Initialize model instance
model = CGDETRPredictor('checkpoint.ckpt',
                        device=device,
                        feature_name='clip')

# Encode video features
model.encode_video('video.mp4')

# Moment retrieval & highlight detection
query = 'A man is speaking in front of the camera'
pred = model.predict(query)
\end{lstlisting}
\end{figure}

\begin{figure*}[t]
    \centering
    \includegraphics[width=0.8\linewidth]{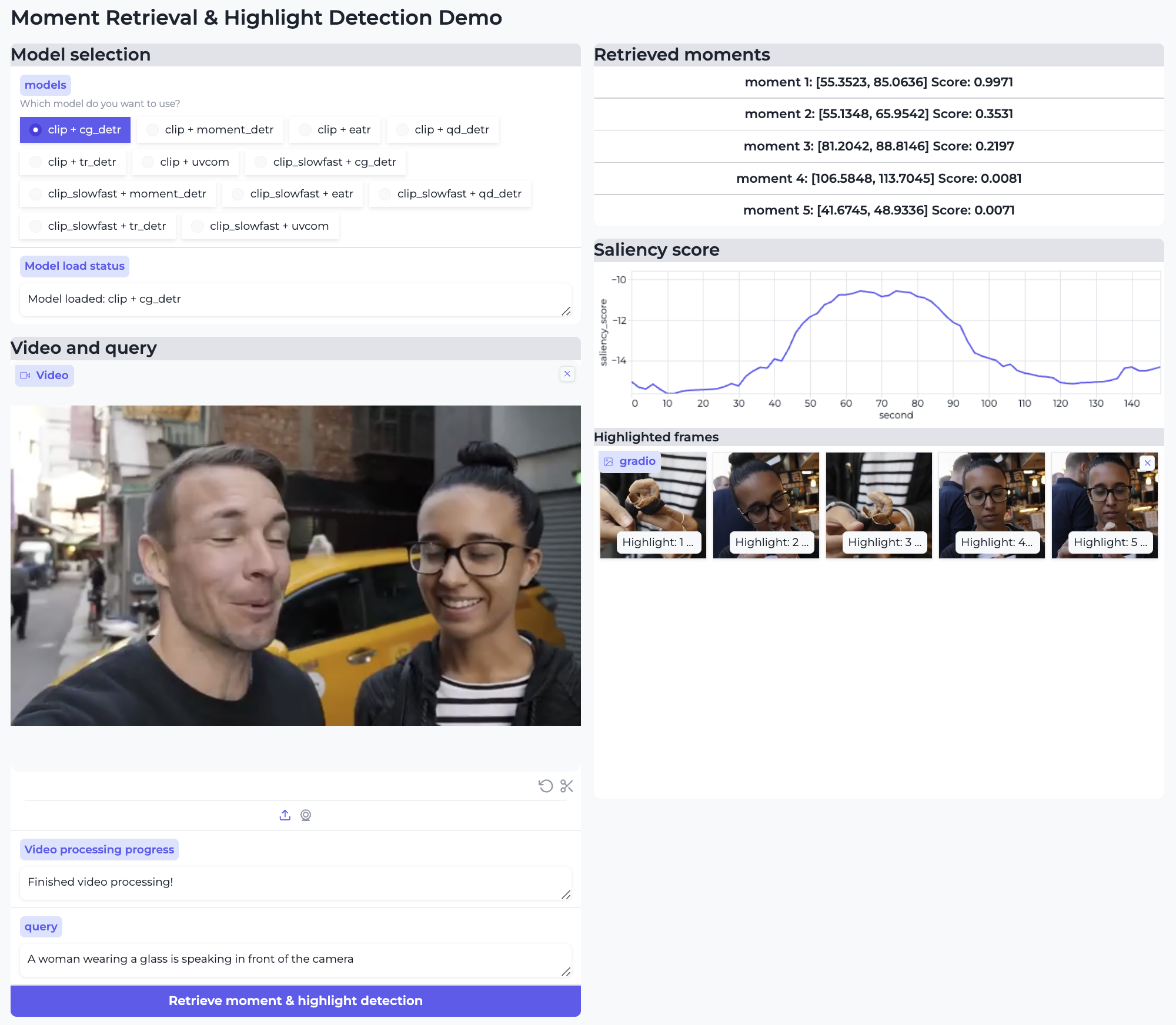}
    \caption{A screenshot of the web demo. In the web demo, you can select a model and feature in the model selection pane. Then, in the video and query pane, you can upload a video and input a text query. By clicking the 'Retrieve Moment \& Highlight Detection' button, the retrieved moments and highlighted frames will be displayed in the right panes. Hugging face spaces: \url{https://huggingface.co/spaces/awkrail/lighthouse_demo}.}
    \label{fig:demo_apps}
\end{figure*}

\noindent\textbf{Easy to use:}
Lighthouse provides an inference API and a web demo, enabling researchers and developers who are not well-versed in detailed MR-HD pipelines, to use MR-HD.
Listing \ref{listing:util} shows the inference API, which hides the detailed implementation of video-text processing and provides users with three main steps: model initialization, \texttt{encode\_video()}, and \texttt{predict()}.
First, the user initializes the model instance by specifying the model weight, device type (i.e., CPU or GPU), and feature name. Second, given a video path, \texttt{encode\_video()} extracts frames from the video, converts them into features, and stores them as instance variables. Finally, given a query, \texttt{predict()} encodes the query and forwards both the video and query features into the model to obtain results. \figref{fig:demo_apps} shows a web demo built upon the inference API to visualize the model's outputs.
By clicking on the moment panes, the video seek bar jumps to the corresponding timestamps, enabling users to view those specific moments. Hovering over the saliency scores lets users see both the values and the corresponding timestamps in the video.

\section{Architecture of Lighthouse}

\begin{figure*}[t]
    \centering
    \includegraphics[width=\linewidth]{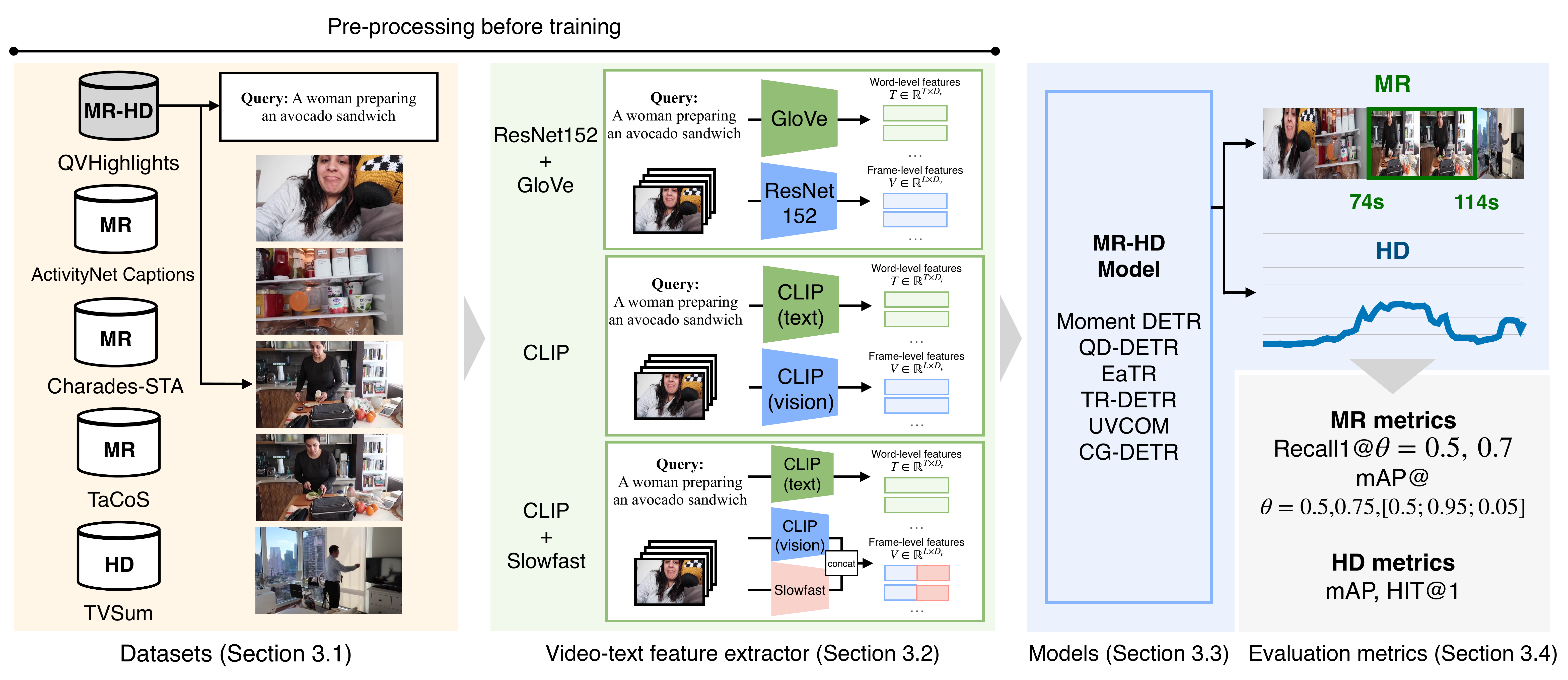}
    \caption{Overview of Lighthouse architecture for MR-HD training and evaluation. It consists of four components: datasets, video-text feature extractor, models, and evaluation metrics.}
    \label{fig:architecture}
\end{figure*}

\figref{fig:architecture} shows an overview of Lighthouse architecture, consisting of four components: datasets, video-text feature extractor, models, and evaluation metrics.

\subsection{Datasets}
\label{sec:dataset}
We utilize five commonly-used datasets: QVHighlights~\cite{lei2021neurips}, ActivityNet Captions \cite{krishna2017iccv}, Charades-STA \cite{hendricks2017iccv}, TaCoS \cite{tacos}, and TVSum \cite{song2015cvpr}. The QVHighlights dataset is an MR-HD dataset comprising videos, queries, and annotations for both moments and highlights.
It is the only dataset that includes annotations for both moments and highlights. Moments are represented as start and end timestamps for each query, while highlights are represented as saliency scores ranging from 1 (very bad) to 5 (very good) for each frame of the video. ActivityNet Captions, Charades-STA, and TaCoS are MR datasets because they contain only moment annotations, whereas TVSum is an HD dataset as it includes 50 videos from ten domains (e.g., news and documentary) and highlight annotations. Note that we do not release the original videos due to copyright issues. Instead, we release the pre-processed video-text features to allow researchers to reproduce experiments.

\subsection{Video-text feature extractor}
\label{sec:feature}
Given video frames and a query, the video-text encoders convert them into frame- and word-level features $\mathbf{V} \in \mathbb{R}^{L\times D_v}, \mathbf{T} \in \mathbb{R}^{T\times D_t}$, where $L$ and $T$ represent the numbers of frames and words, and $D_v$ and $D_t$ represent the dimensions of the vision and text features.
We utilize three feature extractors: CLIP~\cite{radford2021icml}, CLIP+Slowfast~\cite{Feichtenhofer_2019_ICCV}, and ResNet152+GloVe~\cite{he2016cvpr,glove}.
CLIP employs vision and text encoders, based on the Transformer architecture~\cite{attention}, pre-trained on extensive web image-text pairs.
These encoders transform frames and queries into feature vectors.
CLIP+Slowfast combines CLIP vision features with Slowfast features to enhance motion awareness, as Slowfast is pre-trained on the Kinetics400 action recognition dataset~\cite{kinetics} and is adept at recognizing motion in videos. ResNet152+GloVe uses ResNet152 for frame-wise visual features and GloVe for word-level text features. ResNet152 and GloVe are pre-trained on ImageNet~\cite{imagenet} and English Wikipedia, respectively. While CLIP is the standard in MR-HD, this setup allows us to assess the superiority of CLIP's vision-language encoders by comparing them with models trained separately on visual and textual data. Note that we extract video-text features as a preprocessing step before training, rather than during training because extracting features during training is costly and time-consuming. For this process, we use the HERO video extractor library~\cite{hero}.

\begin{table*}[t]
\centering
\scalebox{0.7}{
\begin{tabular}{lcccccccccccccc}
\toprule
\multicolumn{8}{c|}{val} & \multicolumn{7}{c}{test} \\ \midrule
 & \multicolumn{5}{c}{\textbf{MR}} & \multicolumn{2}{c|}{\textbf{HD}} & \multicolumn{5}{c}{\textbf{MR}} & \multicolumn{2}{c}{\textbf{HD}} \\
 & \multicolumn{2}{c}{R1} & \multicolumn{3}{c}{mAP} & \multicolumn{2}{c|}{Very Good} & \multicolumn{2}{c}{R1} & \multicolumn{3}{c}{mAP} & \multicolumn{2}{c}{Very Good} \\ \cmidrule(lr){2-3} \cmidrule(lr){4-6} \cmidrule(lr){7-8} \cmidrule(lr){9-10} \cmidrule(lr){11-13} \cmidrule(lr){14-15}
 & @0.5 & @0.7 & @0.5 & @0.75 & avg & mAP & \multicolumn{1}{c|}{HIT@1} & \multicolumn{1}{l}{@0.5} & \multicolumn{1}{l}{@0.7} & \multicolumn{1}{l}{@0.5} & \multicolumn{1}{l}{@0.75} & \multicolumn{1}{l}{avg} & mAP & HIT@1 \\ \hline
\multicolumn{15}{l}{\textit{\textbf{ResNet152+GloVe}}} \\
\ \ Moment DETR & 41.5 & 25.2 & 45.9 & 22.6 & 24.7 & 29.1 & \multicolumn{1}{c|}{41.4} & 40.0 & 22.0 & 44.9 & 21.6 & 23.8 & 30.0 & 42.9 \\
\ QD-DETR & 53.2 & 37.5 & 55.4 & 34.5 & 34.5 & 34.1 & \multicolumn{1}{c|}{52.1} & 52.7 & 36.1 & 55.4 & 33.9 & 33.7 & 33.8 & 50.7 \\
\ \ EaTR & \textbf{54.9} & 36.0 & \textbf{56.7} & 33.5 & 34.1 & \textbf{35.1} & \multicolumn{1}{c|}{\textbf{54.7}} & \textbf{57.2} & \textbf{38.9} & \textbf{59.6} & \textbf{35.6} & \textbf{36.7} & \textbf{36.3} & \textbf{57.4} \\
\ \ TR-DETR & 48.3 & 32.9 & 49.5 & 28.6 & 29.6 & 34.2 & \multicolumn{1}{c|}{51.4} & 47.7 & 31.6 & 49.8 & 29.3 & 29.4 & 34.3 & 52.0 \\
\ \ UVCOM & 53.7 & \textbf{39.7} & 55.9 & \textbf{36.5} & \textbf{36.1} & 34.9 & \multicolumn{1}{c|}{53.0} & 53.8 & 37.6 & 55.1 & 33.4 & 34.0 & 34.8 & 53.8 \\
\ \ CG-DETR & 51.9 & 39.0 & 54.3 & 36.0 & 35.5 & 34.1 & \multicolumn{1}{c|}{53.2} & 53.1 & 38.3 & 55.7 & 35.1 & 35.1 & 34.5 & 52.9 \\
\multicolumn{15}{l}{\textit{\textbf{CLIP}}} \\
\ \ Moment DETR & 53.5 & 34.1 & 56.2 & 30.8 & 32.4 & 35.3 & \multicolumn{1}{c|}{54.0} & 55.8 & 33.8 & 58.2 & 31.2 & 32.7 & 35.7 & 55.8 \\
\ \ QD-DETR & 59.7 & 42.3 & 60.4 & 37.5 & 37.5 & 38.0 & \multicolumn{1}{c|}{59.2} & 60.8 & 41.8 & 62.3 & 37.1 & 38.3 & 38.2 & 60.7 \\
\ \ EaTR & 54.9 & 36.0 & 56.7 & 33.5 & 34.1 & 35.1 & \multicolumn{1}{c|}{54.7} & 54.6 & 34.0 & 57.1 & 32.6 & 33.2 & 34.9 & 54.7 \\
\ \ TR-DETR & 63.6 & 43.9 & 62.9 & 39.7 & 39.6 & \textbf{40.1} & \multicolumn{1}{c|}{63.2} & 60.2 & 41.4 & 60.1 & 37.0 & 37.2 & 38.6 & 59.3 \\
\ \ UVCOM & 64.8 & 48.0 & 64.2 & 42.7 & 42.3 & 38.7 & \multicolumn{1}{c|}{62.2} & 62.7 & \textbf{46.9} & 63.6 & \textbf{42.6} & \textbf{42.1} & \textbf{39.8} & \textbf{64.5} \\
\ \ CG-DETR & \textbf{66.6} & \textbf{49.9} & \textbf{66.2} & \textbf{44.2} & \textbf{43.9} & 39.9 & \multicolumn{1}{c|}{\textbf{64.3}} & \textbf{64.5} & 46.0 & \textbf{64.8} & 41.6 & 41.8 & 39.4 & 64.3 \\
\multicolumn{15}{l}{\textit{\textbf{CLIP+Slowfast (Reproduced scores)}}} \\
\ \ Moment DETR & 54.2 & 36.1 & 55.3 & 31.5 & 32.6 & 35.9 & \multicolumn{1}{c|}{56.7} & 54.4 & 33.9 & 55.2 & 29.7 & 31.5 & 32.6 & 56.7 \\
\ \ QD-DETR & 63.0 & 46.4 & 63.3 & 41.1 & 41.3 & 39.1 & \multicolumn{1}{c|}{61.3} & 62.1 & 44.6 & 63.0 & 41.0 & 40.6 & 38.8 & 61.6 \\
\ \ EaTR & 59.6 & 40.3 & 60.9 & 38.1 & 38.0 & 36.6 & \multicolumn{1}{c|}{57.9} & 57.2 & 38.9 & 59.6 & 35.6 & 36.7 & 36.6 & 57.9 \\
\ \ TR-DETR & 66.5 & 48.8 & 65.3 & 44.3 & 43.4 & \textbf{40.8} & \multicolumn{1}{c|}{66.2} & \textbf{65.2} & \textbf{48.8} & 64.4 & \textbf{43.0} & 42.6 & 39.8 & 62.1 \\
\ \ UVCOM & 64.0 & 49.4 & 63.3 & 44.8 & 43.9 & 39.7 & \multicolumn{1}{c|}{64.3} & 62.6 & 47.6 & 62.4 & 42.4 & 42.5 & 39.6 & 62.8 \\
\ \ CG-DETR & \textbf{65.6} & \textbf{52.1} & \textbf{65.6} & \textbf{46.3} & \textbf{45.3} & 40.7 & \multicolumn{1}{c|}{\textbf{67.0}} & 64.9 & 48.1 & \textbf{64.8} & 42.8 & \textbf{43.3} & \textbf{40.7} & \textbf{67.0} \\
\midrule
\multicolumn{15}{l}{\textit{\textbf{Reported scores in the reference papers (CLIP+Slowfast)}}} \\
\ \ Moment DETR & 53.9 & 34.8 & - & - & 32.2 & 35.7 & \multicolumn{1}{c|}{55.6} & 52.9 & 33.0 & 54.8 & 29.4 & 30.7 & 35.7 & 55.6 \\
\ \ QD-DETR & 62.7 & 46.7 & 62.2 & 41.8 & 41.2 & 39.1 & \multicolumn{1}{c|}{63.0} & 62.4 & 45.0 & 62.5 & 39.9 & 39.9 & 38.9 & 62.4 \\
\ \ EaTR & 61.4 & 45.8 & 61.9 & 41.9 & 41.7 & 37.2 & \multicolumn{1}{c|}{58.7} & - & - & - & - & - & - & - \\
\ \ TR-DETR & - & - & - & - & - & - & \multicolumn{1}{c|}{-} & 64.6 & \textbf{48.9} & 63.9 & \textbf{43.7} & 42.6 & 39.9 & 63.4 \\
\ \ UVCOM & - & - & - & - & - & - & \multicolumn{1}{c|}{-} & 63.6 & 47.5 & 63.4 & 42.7 & 43.2 & 39.7 & 64.2 \\
\ \ CG-DETR & \textbf{67.4} & \textbf{52.1} & \textbf{65.6} & \textbf{45.7} & \textbf{44.9} & \textbf{40.8} & \multicolumn{1}{c|}{\textbf{66.7}} & \textbf{65.4} & 48.4 & \textbf{64.5} & 42.8 & \textbf{42.9} & \textbf{40.3} & \textbf{66.2} \\ \bottomrule
\end{tabular}
}
\caption{MR-HD results on the QVHighlights dataset. \textbf{Bold} values represent the best scores among methods with the same video-text feature.}
\label{tab:qvhighlight_results}
\end{table*}

\begin{table*}[t]
\centering
\scalebox{0.7}{
\begin{tabular}{lccccc|ccccc|ccccc}
\toprule
& \multicolumn{5}{c|}{ActivityNet Captions} & \multicolumn{5}{c|}{Charades-STA} & \multicolumn{5}{c}{TaCoS} \\ \midrule
& \multicolumn{2}{c}{R1} & \multicolumn{3}{c|}{mAP} & \multicolumn{2}{c}{R1} & \multicolumn{3}{c|}{mAP} & \multicolumn{2}{c}{R1} & \multicolumn{3}{c}{mAP} \\ \cmidrule(lr){2-3} \cmidrule(lr){4-6} \cmidrule(lr){7-8} \cmidrule(lr){7-8} \cmidrule(lr){9-11} \cmidrule(lr){12-13} \cmidrule(lr){14-16}
& @0.5 & @0.7 & @0.5 & @0.75 & avg & @0.5 & @0.7 & @0.5 & @0.75 & avg & @0.5 & @0.7 & @0.5 & @0.75 & avg \\ \midrule
\multicolumn{16}{l}{\textit{\textbf{ResNet152+GloVe}}} \\
\ \ Moment DETR & 34.2 & 19.5 & 46.3 & 24.4 & 26.2 & 38.4 & 22.9 & 52.4 & 22.2 & 26.2 & 20.0 & 8.6 & 24.2 & 6.9 & 10.1 \\
\ \ QD-DETR & 35.4 & 20.3 & 47.4 & 24.9 & 26.6 & \textbf{42.1} & \textbf{24.0} & 56.7 & \textbf{24.5} & \textbf{28.7} & 30.6 & 15.1 & 35.1 & 12.3 & 16.1 \\
\ \ EaTR & 32.4 & 18.2 & 44.3 & 21.9 & 24.1 & 37.6 & 20.1 & 53.5 & 23.6 & 27.0 & 22.5 & 9.2 & 26.3 & 7.9 & 10.7 \\
\ \ UVCOM & 34.4 & 19.9 & 46.1 & 24.4 & 25.9 & 38.1 & 18.2 & 54.4 & 21.1 & 25.6 & 24.1 & 10.7 & 28.1 & 8.6 & 12.0 \\
\ \ CG-DETR & \textbf{37.0} & \textbf{21.2} & \textbf{48.6} & \textbf{26.5} & \textbf{28.0} & 39.7 & 19.4 & \textbf{56.9} & 23.2 & 27.5 & \textbf{34.2} & \textbf{17.4} & \textbf{39.7} & \textbf{14.6} & \textbf{18.7} \\
\multicolumn{16}{l}{\textit{\textbf{CLIP}}} \\
\ \ Moment DETR & 36.1 & 20.4 & 48.2 & 25.7 & 27.5 & 47.9 & 26.7 & 61.0 & 28.8 & 31.9 & 18.0 & 7.9 & 21.3 & 6.7 & 9.3 \\
\ \ QD-DETR & 36.9 & 21.4 & 48.4 & 26.3 & 27.6 & 52.0 & 31.7 & 63.6 & 29.4 & 33.4 & 32.3 & 17.2 & 36.0 & 14.1 & 17.5 \\
\ \ EaTR & 34.6 & 19.7 & 45.1 & 23.1 & 24.9 & 48.4 & 27.5 & 59.9 & 26.9 & 30.9 & 24.7 & 10.0 & 28.8 & 8.7 & 11.8 \\
\ \ UVCOM & 37.0 & 21.5 & 48.3 & 25.7 & 27.4 & 48.4 & 27.1 & 60.9 & 27.9 & 31.4 & \textbf{36.8} & \textbf{20.0} & \textbf{41.5} & \textbf{16.3} & \textbf{20.1} \\
\ \ CG-DETR & \textbf{38.8} & \textbf{22.6} & \textbf{50.6} & \textbf{27.5} & \textbf{28.9} & \textbf{54.4} & \textbf{31.8} & \textbf{65.5} & \textbf{30.5} & \textbf{34.5} & 34.3 & 19.8 & 38.6 & 15.8 & 19.0 \\
\multicolumn{16}{l}{\textit{\textbf{CLIP+Slowfast (Reproduced scores)}}} \\
\ \ Moment DETR & 36.5 & 21.1 & 48.4 & 26.0 & 27.4 & 53.4 & 30.7 & 62.0 & 29.1 & 32.6 & 25.5 & 12.9 & 29.1 & 10.3 & 13.3 \\
\ \ QD-DETR & 37.5 & 22.1 & 48.9 & 26.4 & 27.8 & \textbf{59.4} & \textbf{37.9} & \textbf{66.6} & \textbf{33.8} & \textbf{36.4} & 38.7 & 22.1 & 42.9 & 16.7 & 20.9 \\
\ \ EaTR & 34.6 & 19.3 & 45.2 & 22.3 & 24.6 & 55.2 & 33.1 & 65.4 & 30.4 & 34.2 & 31.7 & 15.6 & 37.4 & 14.0 & 17.2 \\
\ \ UVCOM & 37.3 & 21.6 & 48.9 & 25.7 & 27.3 & 56.9 & 35.9 & 65.6 & 33.6 & 36.2 & 40.2 & 23.3 & 43.5 & 19.1 & 22.1 \\
\ \ CG-DETR & \textbf{40.0} & \textbf{23.2} & \textbf{51.0} & \textbf{27.7} & \textbf{29.2} & 57.6 & 35.1 & 65.9 & 30.9 & 35.0 & \textbf{39.8} & \textbf{25.1} & \textbf{44.2} & \textbf{19.6} & \textbf{22.9} \\
\midrule
\multicolumn{16}{l}{\textit{\textbf{Reported scores in the reference papers (CLIP+Slowfast)}}} \\
\ \ Moment DETR & - & - & - & - & - & 52.1 & 30.6 & - & - & - & 24.7 & 12.0 & - & - & - \\
\ \ QD-DETR & - & - & - & - & - & 57.3 & 32.6 & - & - & - & - & - & - & - & - \\
\ \ EaTR & - & - & - & - & - & - & - & - & - & - & - & - & - & - & - \\
\ \ UVCOM & - & - & - & - & - & \textbf{59.3} & \textbf{36.6} & - & - & - & 36.4 & \textbf{23.3} & - & - & - \\
\ \ CG-DETR & - & - & - & - & - & 58.4 & 36.3 & - & - & - & \textbf{39.6} & 22.2 & - & - & - \\
\bottomrule
\end{tabular}
}
\caption{MR results on the ActivityNet Captions, Charades-STA, and TaCoS datasets.}
\label{tab:charades_results}
\end{table*}

\subsection{Models}
\label{sec:model}

We implement six recent MR-HD models: Moment DETR~\cite{lei2021neurips}, QD-DETR~\cite{Moon_2023_CVPR}, EaTR~\cite{Jang_2023_ICCV}, TR-DETR~\cite{sum2024aaai}, UVCOM~\cite{Xiao_2024_CVPR}, and CG-DETR~\cite{Moon2024arxiv}. These models are extensions of DETR~\cite{detr}, Transformer-based object detectors, adapted for MR-HD. Given a video and language query, they can predict both moments and saliency scores. Note that, except for TR-DETR, these models are designed to be trainable on a single task of MR or HD\footnote{Note that TR-DETR is unavailable for single MR and HD tasks because the official code necessitates MR-HD annotations for loss calculation. See: \url{https://github.com/mingyao1120/TR-DETR/issues/3} for details.}.

We describe briefly by focusing on the difference between these methods.
Moment DETR is first proposed with QVHighlights as an MR-HD baseline. Given video and text features, the Transformer encoder concatenates and encodes them, then the Transformer decoder with query slots predicts both moments and saliency scores. Based on Moment DETR, QD-DETR focuses on enhancing query-moment similarity by introducing contrastive learning using query and different video pairs. EaTR improves Moment DETR by incorporating video and query information into the query slots.
TR-DETR explores the reciprocal relationship between MR and HD to improve performance. UVCOM devises local and global encoding approaches based on the observation that a model shows different attention maps for MR and HD. Specifically, the attention map for MR emphasizes local moments in the videos, whereas, for HD, it highlights a global pattern. CG-DETR also focuses on the attention heatmap between video frames and queries. To achieve this, CG-DETR introduces an adaptive Cross Attention layer, which adds dummy tokens to the key in the multi-head attention to adjust relevancy between words and moments.

\noindent
\textbf{Extension to other model types.} Currently, the models used are based on DETR, and the inference APIs are specifically designed for it. However, research by \cite{mrblip} has shown that the BLIP2-style \cite{blip2} auto-regressive approach outperforms DETR-based models, though it requires significantly more GPU resources (e.g., 8x NVIDIA A100 80GB GPUs for training).
To integrate this into Lighthouse, we believe the frame and video-text feature extraction modules can be shared, and a wrapper class will be needed for the model's forward module. Extending support to other model types is planned for future work.

\subsection{Evaluation metrics}
\label{sec:eval}

We follow the evaluation metrics described in \citet{lei2021neurips}. For MR, we provide Recall1@$\theta$ and mAP@$\theta$. Recall1@$\theta$ represents the percentage of the top 1 retrieved moment with an IoU greater than $\theta$ with the ground-truth moment, where $\theta$ is set to be 0.5 and 0.7. mAP@$\theta$ denotes the mean average precision with $\theta$ set to 0.5 and 0.75, as well as the average mAP across multiple $\theta$ values ranging from 0.5 to 0.95 in increments of 0.05. For HD, we provide mAP, and HIT@1, which computes the hit ratio for the highest scored frame. Note that the frame is regarded as positive if it has a score of ``Very Good ($=5$).'' QVHighlights consists of saliency scores from three annotators, HIT@1 is computed as the average of these annotators.

\section{Experiments}

We perform experiments on MR-HD, MR, and HD tasks individually. We used 1 NVIDIA A100 GPU (48GB) for all experiments. The hyperparameters used in this paper are the same as in the reference papers.

\subsection{MR-HD results}

\tabref{tab:qvhighlight_results} presents MR-HD results on the validation and test splits of QVHighlights, revealing three key insights. First, when comparing the reproduced results using CLIP+Slowfast with the reported scores, Lighthouse generally reproduces the reported scores. The models proposed in 2024, TR-DETR, UVCOM, and CG-DETR, achieve competitive performance among the methods. Second, CLIP+Slowfast generally achieves higher performance than CLIP alone, indicating that sequential motion information in videos is effective for MR-HD tasks in addition to frame-level appearance representations. Finally, CLIP-based features outperform ResNet152+GloVe, demonstrating the effectiveness of CLIP in the MR-HD task.

\subsection{MR results}

\tabref{tab:charades_results} presents the MR results. Although the insights gained are similar to the MR-HD results, we observe one different finding; later methods do not consistently outperform older ones across different datasets and features. For instance, in Charades-STA, QD-DETR with CLIP+Slowfast and ResNet152+GloVe achieves higher performance than CG-DETR and UVCOM. This suggests that there is no one-size-fits-all solution. To apply the methods to a custom MR dataset, users need to test multiple methods with different features. Lighthouse facilitates this trial-and-error process to achieve the best performance settings.

\subsection{HD results}

\begin{table}[t]
\centering
\scalebox{0.48}{
\begin{tabular}{llllllllllll}
\toprule
 & VT & VU & GA & MS & PK & PR & FM & BK & BT & DS & avg \\
\midrule
\multicolumn{12}{l}{\textit{\textbf{ResNet152+GloVe}}} \\
\ \ Moment DETR & 87.5 & \textbf{93.3} & 91.5 & 79.7 & \textbf{92.6} & 85.1 & 70.0 & 91.8 & 87.9 & 79.7 & 85.9 \\
\ \ QD-DETR & \textbf{90.8} & 89.8 & 90.8 & 83.6 & 88.8 & 85.3 & \textbf{79.6} & \textbf{95.1} & 89.7 & 78.4 & 87.2 \\
\ \ EaTR & 87.9 & 87.2 & 89.0 & \textbf{87.9} & 85.8 & 90.1 & 73.2 & 92.3 & 89.4 & 78.7 & 86.2 \\
\ \ UVCOM & 87.8 & 92.6 & \textbf{94.7} & 80.7 & 88.7 & \textbf{91.3} & 76.0 & 94.0 & 90.1 & \textbf{80.1} & \textbf{87.6} \\
\ \ CG-DETR & 89.3 & 89.3 & 93.6 & 84.8 & 89.5 & 86.5 & 76.4 & 93.6 & \textbf{90.2} & 77.9 & 87.1 \\
\multicolumn{12}{l}{\textit{\textbf{CLIP}}} \\
\ \ Moment DETR & \textbf{92.0} & \textbf{95.8} & \textbf{96.5} & 87.3 & 89.0 & 89.9 & \textbf{80.4} & 92.6 & 87.8 & 79.5 & \textbf{89.1} \\
\ \ QD-DETR & 88.5 & 92.6 & 94.4 & 86.2 & 88.0 & \textbf{91.9} & 78.6 & 94.0 & 90.0 & 79.6 & 88.4 \\
\ \ EaTR & 86.4 & 94.1 & 90.9 & 84.9 & 83.8 & 88.9 & 77.9 & 92.5 & \textbf{90.8} & 76.8 & 86.7 \\
\ \ UVCOM & 90.1 & 92.4 & 95.8 & 86.5 & 86.8 & 89.2 & 76.5 & \textbf{95.4} & 87.7 & 76.1 & 87.7 \\
\ \ CG-DETR & 89.7 & 86.3 & 91.0 & \textbf{90.6} & \textbf{90.6} & 89.4 & 75.4 & 95.1 & 90.0 & \textbf{83.2} & 88.1 \\
\multicolumn{12}{l}{\textit{\textbf{CLIP+Slowfast}}} \\
\ \ Moment DETR & 85.0 & \textbf{95.8} & 91.6 & 88.2 & 85.8 & 85.2 & 76.3 & 91.8 & 88.0 & 81.3 & 86.9 \\
\ \ QD-DETR & \textbf{90.3} & 93.2 & 91.3 & 85.0 & \textbf{90.9} & 88.9 & \textbf{78.6} & 94.0 & 88.7 & \textbf{82.9} & 88.4 \\
\ \ EaTR & 87.1 & 93.7 & 89.5 & 84.6 & 88.5 & 84.5 & 73.4 & 91.4 & 88.8 & 79.9 & 86.1 \\
\ \ UVCOM & 89.6 & 92.8 & 91.4 & 87.4 & 87.9 & 86.9 & 76.3 & \textbf{95.4} & \textbf{90.2} & 79.5 & 87.7 \\
\ \ CG-DETR & 89.0 & 92.6 & \textbf{96.3} & \textbf{92.0} & 88.9 & \textbf{89.2} & 77.0 & 94.0 & 87.4 & 81.9 & \textbf{88.8} \\
\multicolumn{12}{l}{\textit{\textbf{I3D+CLIP (Text) (Reproduced scores)}}} \\
\ \ Moment DETR & 84.6 & 93.5 & 91.7 & 80.8 & 88.4 & \textbf{91.4} & 77.3 & 92.5 & 88.6 & 78.1 & 86.7 \\
\ \ QD-DETR & 89.9 & 86.6 & 91.1 & 85.9 & 88.7 & 88.9 & 74.2 & \textbf{97.1} & 88.3 & 80.0 & 87.1 \\
\ \ EaTR & 86.9 & 80.3 & 91.4 & 75.2 & 88.9 & 86.1 & 76.8 & 93.1 & 88.6 & \textbf{82.5} & 85.0 \\
\ \ UVCOM & 89.2 & \textbf{92.4} & \textbf{94.4} & 91.1 & 84.4 & 89.9 & \textbf{77.8} & 94.0 & 87.3 & 78.8 & \textbf{87.9} \\
\ \ CG-DETR & \textbf{90.5} & 83.1 & 94.2 & \textbf{91.9} & \textbf{90.6} & 88.6 & 76.1 & 94.0 & \textbf{89.1} & 81.0 & \textbf{87.9} \\
\midrule
\multicolumn{12}{l}{\textit{\textbf{Reported scores in the reference papers (I3D+CLIP (Text))}}} \\
\ \ Moment DETR & - & - & - & - & - & - & - & - & - & - & - \\
\ \ QD-DETR & \textbf{88.2} & 87.4 & 85.6 & 85.0 & 85.8 & 86.9 & 76.4 & 91.3 & \textbf{89.2} & 73.7 & 85.0 \\
\ \ EaTR & - & - & - & - & - & - & - & - & - & - & - \\
\ \ UVCOM & 87.6 & \textbf{91.6} & 91.4 & 86.7 & \textbf{86.9} & 86.9 & \textbf{76.9} & 92.3 & 87.4 & 75.6 & 86.3 \\
\ \ CG-DETR & 86.9 & 88.8 & 94.8 & \textbf{87.7} & 86.7 & \textbf{89.6} & 74.8 & \textbf{93.3} & \textbf{89.2} & \textbf{75.9} & \textbf{86.8} \\ 
\bottomrule
\end{tabular}
}
\caption{HD results on TVSum. mAP scores for each domain are displayed.}
\label{tab:tvsum_results}
\end{table}

\tabref{tab:tvsum_results} presents the HD results on the TVSum dataset. In addition to our three backbones, we tested I3D+CLIP (Text) because previous studies used I3D \cite{carreira2017cvpr} and CLIP as visual and textual backbones. The findings are consistent with the MR results. First, the results demonstrate that Lighthouse can reproduce the reported scores. Second, we observe that newer methods do not always outperform older ones across different features. For example, when using CLIP, Moment DETR outperforms other approaches. Thus, Lighthouse is valuable for the HD community to test multiple methods with various features.

\section{Conclusion}
In this paper, we proposed Lighthouse, a user-friendly library for reproducible MR-HD. It supports six methods, five datasets, and three features. Lighthouse includes the inference API and web demo, enabling users to try MR-HD methods easily. Our experiments showed that Lighthouse reproduces the reported scores. In addition, we found that newer MR-HD methods do not consistently outperform older ones across MR/HD datasets and various features. Lighthouse aids researchers in the trial-and-error process, helping them achieve optimal performance settings.

\section{Limitation and future work}
This paper has two main limitations. First, we did not conduct a usability study to assess how the developed demos assist end users. We plan to address this in future work. Second, our models are based on DETR, and we did not implement other types of models. Recently, autoregressive approaches have been introduced in MR \cite{mrblip} based on large language models \cite{t5}.
One of our future directions is to enhance Lighthouse by incorporating these approaches.

\bibliography{custom}

\end{document}